\documentclass{article}
\usepackage{amsmath,epsfig}
\usepackage[preprint]{spconfa4}
\usepackage{xcolor}
\usepackage{cite}
\usepackage{booktabs} 

\let\OLDthebibliography\thebibliography
\renewcommand\thebibliography[1]{
  \OLDthebibliography{#1}
  \setlength{\parskip}{0pt}
  \setlength{\itemsep}{0pt plus 0.3ex}
}

\begin{document}\sloppy

\def\x{{\mathbf x}}
\def\L{{\cal L}}

\title{Migrating Face Swap to Mobile Devices: A lightweight Framework and A Supervised Training Solution}
%
%

\name{Haiming Yu$^{\ast}$, 
Hao Zhu$^{\dagger}$, 
Xiangju Lu$^{\ast \ddagger}$\thanks{$^{\ddagger}$Corresponding author},
Junhui Liu$^{\ast}$}

\address{$^{\ast}$iQIYI Inc, Beijing, China. \{yuhaiming, luxiangju, liujunhui\}@qiyi.com
\\ $^{\dagger}$Nanjing University, Nanjing, China. zhuhaoese@nju.edu.cn}

\maketitle

\begin{abstract}
Existing face swap methods rely heavily on large-scale networks for adequate capacity to generate visually plausible results, which inhibits its applications on resource-constraint platforms. In this work, we propose MobileFSGAN, a novel lightweight GAN for face swap that can run on mobile devices with much fewer parameters while achieving competitive performance. A lightweight encoder-decoder structure is designed especially for image synthesis tasks, which is only 10.2MB and can run on mobile devices at a real-time speed. To tackle the unstability of training such a small network, we construct the FSTriplets dataset utilizing facial attribute editing techniques. FSTriplets provides source-target-result training triplets, yielding pixel-level labels thus for the first time making the training process supervised. We also designed multi-scale gradient losses for efficient back-propagation, resulting in faster and better convergence. Experimental results show that our model reaches comparable performance towards state-of-the-art methods, while significantly reducing the number of network parameters. Codes and the dataset have been released\footnote{https://github.com/HoiM/MobileFSGAN}.
\end{abstract}

\section{Introduction}

Face swap aims at replacing the identity of the face in a target image with that of a source image. This technology has wide applications in various areas including video entertainment, film industry, privacy protection, AR/VR, etc. With the advances in deep learning, many GAN-based methods \cite{fsgan} \cite{tfvgan} \cite{hififace} \cite{faceshifter} \cite{simswap} \cite{megapixels} have been proposed to solve this problem, which significantly improves the practicability compared with traditional industrial-level face swap methods. 

Existing face swap methods, utilizing either 3D modeling or GANs, rely heavily on large-size neural networks to synthesize photo-realistic results, which inevitably limits its applications on resource-constraint platforms. In this paper, we propose MobileFSGAN, a mobile-level face swap framework based on generative adversarial networks. The model size of MobileFSGAN in the inference phase is only $10.2$MB, which is less than $1\%$ of the state-of-the-art models. To the best of our knowledge, MobileFSGAN is the first face swap model that can run on mobile devices. We believe that MobileFSGAN will facilitate the application of face swap especially on mobile phones and AR/VR devices.

Training such a lightweight network is not trivial. Firstly, unlike many other computer vision tasks with ground truth labels for fully-supervised training, it is hard to obtain the face-swapped images for a pixel-wise supervision. Besides, adversarial learning introduces more unstability to the training stage. To mitigate these challenges, we propose FSTriplets, a large-scale face swap dataset that provides source-target-result triplets for training. We observe that the only difference between the result and the target within an ideal triplet is the identity. Therefore, given two face images of the same person, we firstly construct the source-result pair, then we modify the identity of the result image. The modified image will then serve as target. Thus a training triplet is constructed. Providing 'labels' makes the unsupervised training fully supervised and results in a faster and better convergence.

Moreover, we also propose the multi-scale-gradient loss. This mechanism allows supervision not only on the final results, but also on low-resolution outputs produced by the intermediate layers. By enabling gradient flows through intermediate layers, the lightweight generator can be trained efficiently and effectively.

Our contributions are summarized as three aspects:
\vspace{-0.1in}
\begin{itemize}
\item We present MobileFSGAN, the first mobile-level face swap model that can run on mobile devices, reducing the parameters to $1\%$ while achieving competitive performance compared with state-of-the-art methods.
\vspace{-0.1in}
\item We present FSTriplets, a large-scale dataset for face swap that provides ground-truth images to stabilize the training process and enhance the fidelity of the generated images. 
\vspace{-0.1in}
\item We propose the multi-scale-gradient loss that guarantees efficient back-propagation, accelerating and stabilizing the training process.
\end{itemize}
\vspace{-0.1in}

\begin{figure*}[h]
\includegraphics[width=1.0\linewidth]{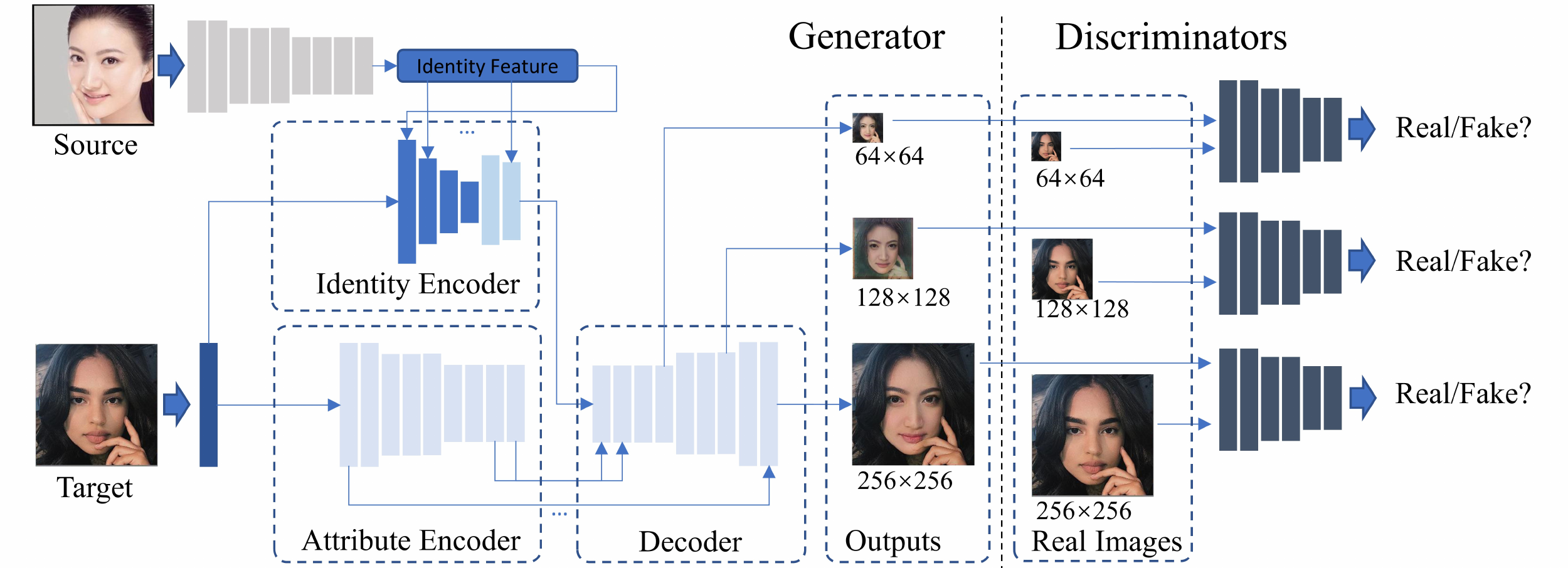}
\centering
\vspace{-0.33in}
\caption{Overall architecture of our network. 
}
\vspace{-0.2in}
\label{fig:overall}
\end{figure*}

\section{Related Works}

\textbf{Face Swap Frameworks}: Recent works on face swap can be classified into two categories: Modeling-based methods \cite{fsgan,tfvgan,hififace} and GAN-based methods \cite{ipgan, faceshifter, simswap}. Since 3DMM \cite{3dmm} provides an explicable approach to the disentanglement of identity and expression, a swapped face can be easily obtained by combining the parameters of the source identity and the target expression. On the other hand, GAN-based methods are another branch of face swap. They inject the source identity information into the generator, implicitly disentangle identities and expressions (e.g. using attention mechanisms). We follow the GAN-based track and propose the first mobile-level framework.


\textbf{Lightweight CNN}: Existing works (e.g. \cite{mobilenet, mobilenet2, squeezenet, shufflenet}) focus mainly on backbone networks for classification, segmentation and detection tasks. These tasks differ from image synthesis that essentially extracts required features from the image data and discard task-irrelevant information. Essentially, these tasks only requires an encoder network. On the other hand, image synthesis requires a decoder to produce visually plausible images thus much more information needs to be preserved. Training such a lightweight model can be even harder without pixel-level supervision.

\section{Approach}

\subsection{Network Design}

MobileFSGAN is a GAN-based framework. The overall architecture is shown in figure \ref{fig:overall}. The generator is an encoder-decoder structure with two encoders and one decoder. The identity encoder accepts the target image as the input, using AdaIN \cite{adain} to inject the source face embedding into the model and replace the target identity, which is formulated as:

\[f_{out} = \frac{f_{in} - \mu}{\sigma} \times \sigma_{id} + \mu_{id}.\]

The input feature maps $f_{in}$ are firstly normalized with the accumulated mean $\mu$ and standard deviation $\sigma$, and then demodulated by the channel-wise learned mean $\mu_{id}$ and standard deviation $\sigma_{id}$ that are conditioned on the source face embedding.

To make the generator lightweight, we firstly fix the number of channels for all feature maps throughout the generator, while at the same time increasing the depth to obtain adequate capacity. This avoids the exponential growth of parameters as the network goes deeper. Meanwhile, feature fusion \cite{densenet} and skip connections \cite{resnet} between the encoder and decoder are deployed, which not only enriches the feature representation but also facilitates the gradient flow. Furthermore, we only downscale the feature maps three times within the encoder, aiming at better preserving the spatial structure. All these efforts result in a generator with its size being only 10.2MB. 

Since AdaIN re-normalizes the feature maps equally in the spatial domain, attribute information residing in the non-face region can be destroyed. Therefore, we also design an attribute encoder to preserve any information that is destroyed by AdaINs throughout the identity encoder. Finally, feature maps from both encoder branches are fused spatially inside the decoder via learned attention mechanisms, preserving the attributes and meanwhile transferring the identity. Readers can refer to the supplementary materials for the detailed structure of our network and corresponding ablation studies. 

\begin{figure*}[h]
\includegraphics[width=0.80\linewidth]{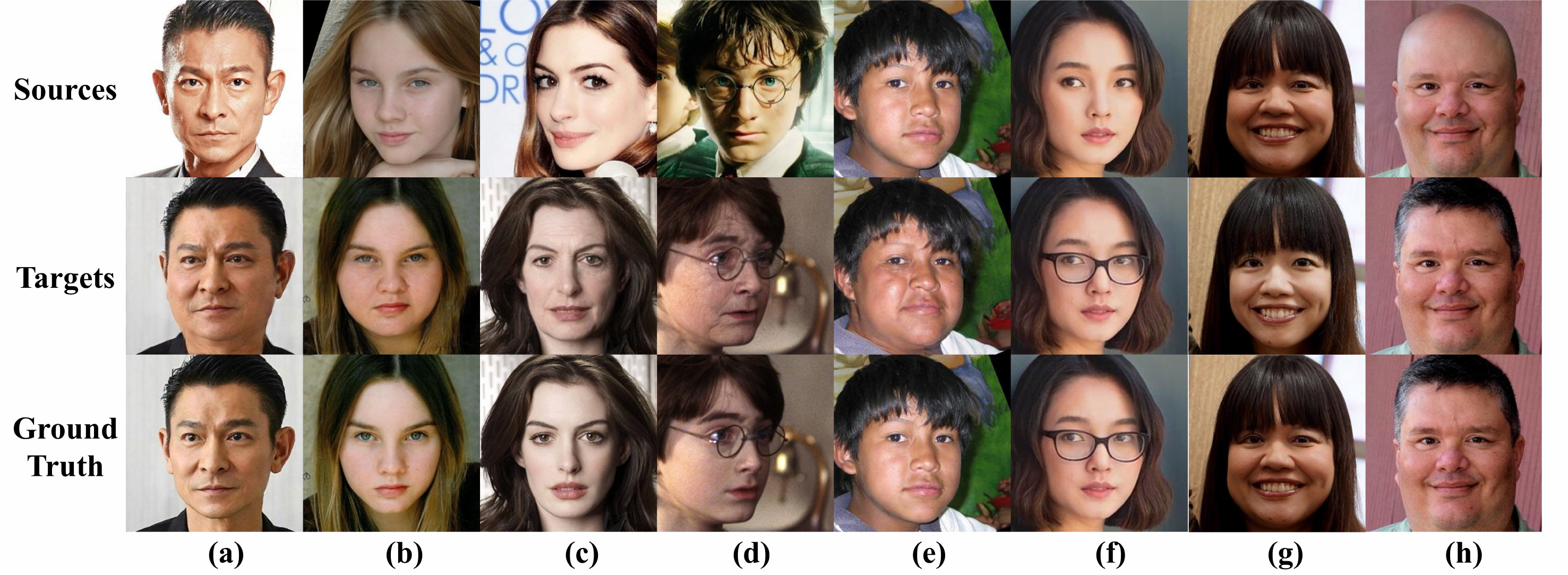}
\centering
\vspace{-0.15in}
\caption{Example triplets in our FSTriplets dataset. (a-d) are constructed with two different images with the same identity; (e-h) illustrates the four types of triplets constructed using a single image. Specifically, identity-related modifications are applied on the target of (e) and the source of (g). Attribute-related modifications are applied on the target of (f) and the source of (h). }
\vspace{-0.2in}
\label{fig:dataset}
\end{figure*}

\subsection{Training Losses}

\textbf{Adversarial Loss}: To improve the image quality, we exploit adversarial training. Specifically, the hinge loss formulation \cite{hingelossgan1} \cite{hingelossgan2} is deployed:

\begin{align*}
L_D = & - E_{x\sim P_{real}}[min(0, -1+D(x))] \\
      & - E_{x\sim P_{fake}}[min(0, -1-D(x))] \\
L_G = & -E[D(G(x_{source}, x_{target}))]
\end{align*}

\textbf{Identity Preserving Loss}: We guarantee identity coherence between the source and the result by forcing large cosine similarity. ArcFace \cite{arcface} is employed to extract face embeddings. Specifically, the MobileNet \cite{mobilenet}-based implementation is used:
\[L_{id} = 1 - cos(f_{src}, f_{res}),\]
where $f_{src}$ and $f_{res}$ are the normalized face embeddings of the source and result images. 

\textbf{Attribute Loss}: 
We use VGG loss to preserve the attributes while allowing modification on the identity. 
\[L_{vgg} = |F(X_{res}) - F(X_{tgt})|,\]
where $F(\cdot)$ is the VGG19 \cite{vgg} feature extractor and $X_{tgt}$ is the target image. Unlike FaceShifter \cite{faceshifter}, we do not need an extra attribute encoder for inference. 

\textbf{Pixel-level Loss}: Finally, mean square error is calculated on the generated image and the ground truth to serve as the pixel-level loss ($L_{pix}$). This provides pixel-level supervision.  

The total loss for training the generator is hence the sum of the above losses: 
\[\ell = \lambda_{G}L_{G} + \lambda_{id}L_{id} + \lambda_{vgg}L_{vgg} + \lambda_{pix}L_{pix} .\]

\textbf{Multi-scale Gradient Losses}
To efficiently update the parameters as well as stabilize the training, we designed multi-scale gradient losses. Specifically, every time before the size of the feature maps doubles in the decoder, we append a convolution layer to produce an RGB image. There are in total three outputs with their sizes being $64\times 64$, $128\times 128$ and $256\times 256$ respectively. During training, we apply adversarial losses and identity-preserving losses on three outputs. We will show how this accelerates the convergence and leads to a better optimal solution.

\subsection{FSTriplets Dataset}


We construct our FSTriplets dataset by editing face images on either the identity or other facial attributes.
Specifically, identity-based modifications include making the face chubby and aging the face, whereas attribute-related modifications include adding glasses, changing hairstyles, making the face grin, etc\cite{stargan, spade}.

Generally, FSTriplets contain over 1.1M images, inside which, the unedited images take up 277.1K. The dataset is constituted with a diverse identity distribution of 15,552 individuals. Each identity contains at least 5 different images and the average amount of images within each identity is 19.13. This enables us to construct an exponential quantity of triplets. Next, we describe two ways to generate triplets. 

\noindent\textbf{Generate triplets from two images of the same person.}
We observe that the only difference between the face-swap result and the target image is the identity. Meanwhile, the result should share the same identity with the source image. When constructing triplets with two different images of the same person, we treat one as the source and the other as the swapped face. We modify the identity of the ground truth to generate the target image. Hence, when we swap the identity inside this target image, the result should be the preset ground truth.

Since face outline shapes and skin textures are typical clues for recognizing identities, we change the face outline shape by making the face chubby and modify the texture by aging the face. Previous works often ignore the face shape by merging the generated face into the target face outline via post-processing. We construct such triplets to provide explicit signals to force face outline shape transfer (see figure \ref{fig:dataset} (a) and (b)). Meanwhile, modifying the texture is beneficial to cases where the source face has great contrast with the target, e.g. swapping across genders or those with a large age span. Examples can be found in (c) and (d) of figure \ref{fig:dataset}.

\noindent\textbf{Generate data from a single image.}
Previous works often use the same image to serve as both the source and the target. Hence the swapped result should also be this image and pixel-level supervision is thus available. However, the portion of such training samples within the whole dataset is always controlled very small in order to prevent overfitting: the output of the generator may copy the target image. However, decreasing pixel-level supervision inevitably brings instability in training.

To prevent overfitting while at the same time enhancing the stability, we also construct triplets with one single image. We can conduct either identity-related or attribute-related modifications on either the source or the target image, which results in four different types of triplets. However, when we conduct identity-related modification, the ground-truth should always be identical to the source image (either modified or not). Meanwhile, if the modification is attribute-related, the ground-truth should always be identical to the target image (either modified or not). Readers can refer to the right four examples of figure \ref{fig:dataset} for a clearer illustration. 

Therefore, the correct result can be identical to either the target or the source, introducing a new regularization thus correctly leading the learning process. This not only guarantees stability but avoids overfitting as well. 

\begin{figure*}[h!bt]
\includegraphics[width=1.0\linewidth]{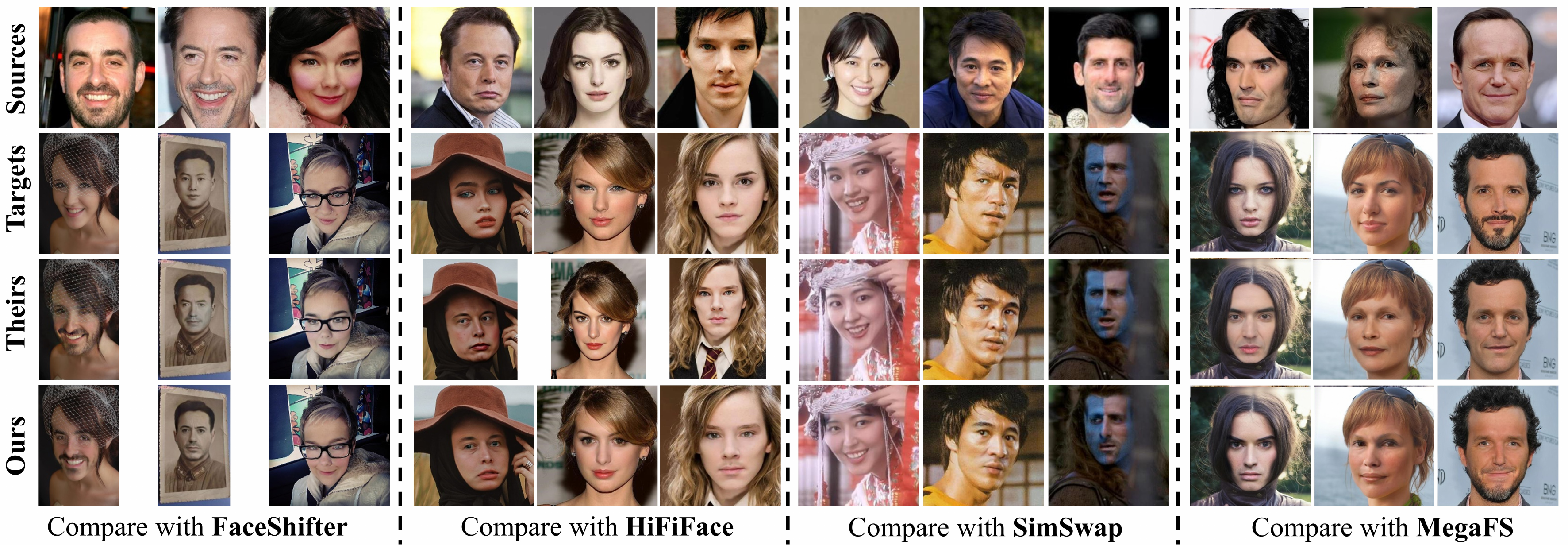}
\centering
\vspace{-0.33in}
\caption{Qualitative comparisons with four SOTA methods. We recommend watching the supplementary videos for more results.}
\vspace{-0.15in}
\label{fig:mixed_comparison}
\end{figure*}

\section{Experiments}



\subsection{Qualitative Results}

We compare MobileFSGAN with four SOTA face swap methods: FaceShifter \cite{faceshifter}, SimSwap \cite{simswap}, HiFiFace \cite{hififace} and MegaFS \cite{megapixels}. As many of these methods do not release their model, we use the results provided by their authors for comparison. The qualitative evaluations are shown in figure \ref{fig:mixed_comparison}. We notice that our model can effectively transfer the source identity, while target expression is well preserved. Moreover, our model is also robust to occlusions, capable of keeping the makeup, and better preserves the face shape of the source images. We recommend watching the demo videos in the supplementary material, which shows our method and model deployed on the mobile device.

\subsection{Quantitative Statistics}

Following the work of FaceShifter \cite{faceshifter}, the experiment is conducted on the FaceForensics++ \cite{faceforensicspp} dataset. 10K face images are generated by evenly sampling 10 frames from each of the 1000 video clips. We then conduct face swap based on the predefined source-target pairs. Three metrics are reported: \textit{ID retrieval}, \textit{pose error} and \textit{expression error}. 

We extract the face embedding of each face using the face recognition model CosineFace \cite{cosineface} and use the cosine similarity to measure the identity distance. For each swapped face, we retrieve the face that has the least identity distance from all frames. We report the retrieval accuracy as the \textit{ID Retrival} metric, which measures identity transferability. We use a pose estimator \cite{deepheadpose} to estimate the head pose of the swapped face. We also use a 3D reconstruction model \cite{accurate3d} to estimate the 3DMM expression parameters. The $\mathcal{L}$-2 distances between the pose vectors and expression vectors of the target image and the swapped image are respectively reported as the \textit{pose} and \textit{expression} metrics to measure the attribute preservation ability. All these statistics are listed in table \ref{tab:faceforensicspp}. 

Our model performs on par with the best-performing model on \textit{ID retrieval}. For \textit{pose error} and \textit{expression error}, our model is also comparable with other SOTA models.  

\begin{table}[tbp]
\centering
\vspace{-0.1in}
\caption{Quantitative comparison on FaceForensics++. 
}
\label{tab:faceforensicspp}
\begin{tabular}{lccc}
\toprule
Method      & ID Retrieval   & Pose          & Expression    \\ \midrule
DeepFake    & 81.96          & 4.14          & -             \\
FaceSwap    & 54.19          & 2.51          & -             \\
IPGAN\cite{ipgan}       & 82.41          & 4.04          & -             \\
FaceShifter\cite{faceshifter} & 97.38          & 2.96          & -             \\
MegaFS\cite{megapixels}      & 82.66          & 4.71          & 3.52          \\
SimSwap\cite{simswap}     & 92.83          & \textbf{1.53} & \textbf{2.56} \\
HiFiFace\cite{hififace}    & \textbf{98.48} & 2.63          & -             \\ 
Ours        & \textbf{98.48} & 2.18          & 2.68     \\ \bottomrule
\end{tabular}
\vspace{-0.1in}
\end{table}

We also validate the identity-transferring ability by conducting experiments on the Labeled Face In-the-wild (LFW) dataset \cite{lfw}, which is often used for face verification. LFW verification set provides a total of $6000$ pairs of profile pictures, among which $3000$ pairs each consist of two different people. In our experiment, we use three accessible face swap models to transfer the identity of the second image to the first image inside every of the $3000$ pairs. Then we use the face-swapped image to calculate the cosine similarity with the second image, since after the swap two images should ideally share the same identity. Example swapped image pairs can be found in the appendix. We list the statistics of the cosine similarity in table \ref{tab:lfw}. Note that the face net used for training is NOT deployed here. We instead use the ResNet50-based ArcFace \cite{arcface} model. Experimental results show that our model is competitive with existing state-of-the-art methods.

\begin{table}[t]
\centering
\caption{Evaluation of similarities and subjective scores.}
\label{tab:lfw}
\begin{tabular}{lccccc}
\toprule
Method  & Mean & Var. & Real & Identity & Attribute \\ \midrule
MegaFS\cite{megapixels}  & 0.5124 & 0.0155 & 2.49 &2.18 & 2.01 \\
SimSwap\cite{simswap} & \textbf{0.6539} & \textbf{0.0043} & 3.31 & 2.59 & \textbf{3.81} \\
Ours    & 0.5920 & 0.0068 & \textbf{3.38} & \textbf{2.77} & 3.53 \\ \bottomrule
\end{tabular}
\vspace{-0.1in}
\end{table}

\subsection{Subjective Evaluation}

We conduct three user studies to further evaluate the performance of our model. Given the source and target images as well as results generated by MegaFS, SimSwap and our model, the users are asked to rate each of the images from three perspectives: i) how realistic the generated images are, ii) how visually similar the identities between the generated and the source images are, iii) how visually similar the attributes (e.g. expression, pose, background, etc.) between the generated and the target images are. Such image sets are randomly sampled from the FaceForensics++ experiments. Valid scores range from 0 to 5 where 5 is the highest score. We collected scores from 18 naive evaluators who were each assigned 40 image sets. The average scores are presented in right three columns of table \ref{tab:lfw}, showing that our model outperforms MegaFS and performs on par with SimSwap.


\subsection{Mobile-level Performance}

Since our model aims at running on end devices, in this section, we validate the mobile-level performance of our model. Firstly, we list the statistics of the inference speed on different platforms. We converted the model under different inference frameworks and test the time for a forward pass on various devices. For each test, we run the forward pass for $100$ times and calculate the average processing time. Note that the time used does not include that of the forward pass of the face net, since the face embedding is only extracted once before face swap. The test results are listed in table \ref{tab:table_speed}.

\begin{table}[h!bt]
\centering
\caption{Inference speed across various platforms, operating systems, inference frameworks, and computing devices. Speed is measured in milliseconds.}
\label{tab:table_speed}
\begin{tabular}{lcccc}
\toprule
Platform                                                      & OS      & Framework & Hardware                                                  & Speed \\ \midrule
iPhone 8                                                      & iOS     & TNN       & GPU                                                       & 398.8 \\
iPhone 11                                                     & iOS     & TNN       & GPU                                                       & 242.4 \\
iPhone 11                                                     & iOS     & CoreML    & ANE                                                       & 135.0 \\
\begin{tabular}[c]{@{}l@{}}Mate 40 Pro\end{tabular} & Android & TNN       & GPU                                                       & 152.5 \\
\begin{tabular}[c]{@{}l@{}}Mate 40 Pro\end{tabular} & Android & TNN       & NPU                                                       & 229.1 \\
PC                                                            & Ubuntu  & PyTorch   & \begin{tabular}[c]{@{}c@{}}1060Ti\end{tabular} & 31.5  \\ \bottomrule
\end{tabular}
\end{table}



Lastly, we compare the size of our model with other existing face swap models. Comparison is illustrated in table \ref{tab:model_size}. To the best of our knowledge, our work is the first lightweight face swap model. The size of our model is much smaller than other previous models, which allows our model to be easily embedded into mobile phone apps and other resource-limited devices. 

\begin{table}[]
\centering
\vspace{-0.2in}
\caption{Comparison of model sizes. 
}
\label{tab:model_size}
\begin{tabular}{lc}
\toprule
Model                 & Model Size (MB)        \\ \midrule
FaceShifter\cite{faceshifter} (PyTorch) & 669.0 + 64.0 + 168.0          \\
MegaFS\cite{megapixels} (PyTorch)      & 3529.0 + 364.0 \\ 
SimSwap\cite{simswap} (PyTorch)      & 766.9 + 220.2 \\ 
Ours (Pytorch)        & 10.2 + 4.9             \\ 
Ours (CoreML)         & 15.0                   \\ 
Ours (TNN)            & 14.8                   \\ \bottomrule

\end{tabular}
\end{table}

\subsection{Ablation Study}

\begin{figure}[t]
\includegraphics[width=1.0\linewidth]{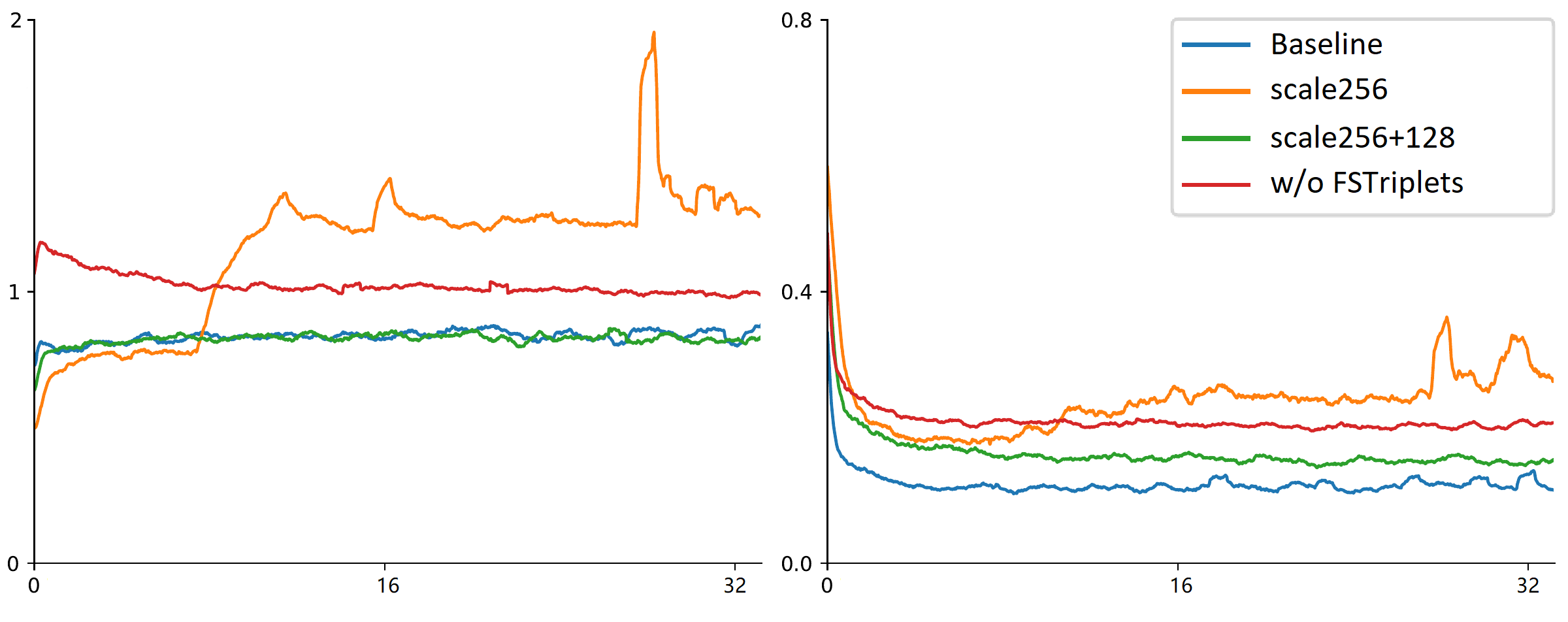}
\centering
\vspace{-0.25in}
\caption{Left: loss curves of the attribute loss. Right: loss curves of the identity preserving loss. Best viewed in color. }
\label{fig:loss_curves}
\vspace{-0.25in}
\end{figure}

To demonstrate the effectiveness of our proposed dataset, we also trained MobileFSGAN without FSTriplets while keeping other settings unchanged. Following the conventional practice, we arrange the training data in the same way as \cite{faceshifter}. Figure \ref{fig:loss_curves} shows the loss curves of the identity preserving loss and the attribute loss calculated on the validation set during training. We can notice that by using our dataset, the optimization process converges faster and leads to a better solution.

We also conducted experiments to show the effectiveness of our multi-scale gradient losses. We trained two extra models by gradually removing intermediate outputs. As can be shown in figure \ref{fig:loss_curves}, training without intermediate outputs (scale256) failed to converge. Meanwhile, training without the $64\times 64$ output (scale256+128) converges worse than our baseline model.

\section{Conclusion}

In this paper, we proposed the first lightweight GAN-based framework for face swap on mobile devices.
To make the training process of such a small model easier, we proposed the FSTriplets dataset, which for the first time makes it possible to train face swap models in a supervised manner. Meanwhile, we designed multi-scale gradient losses to realize efficient parameter updates, resulting in faster and better convergence. Dense experiments and comparisons have demonstrated the effectiveness of our method, which performs competitively well compared with SOTA large-scale models. We hope our research can benefit the application of deep learning on mobile and wearable devices.

\bibliographystyle{IEEEbib}
\bibliography{icme2021template}

\end{document}


\maketitle

\section{Appendix I}

In this section, we provide supplementary qualitative results.

\begin{figure*}[h!bt]
\includegraphics[width=19cm]{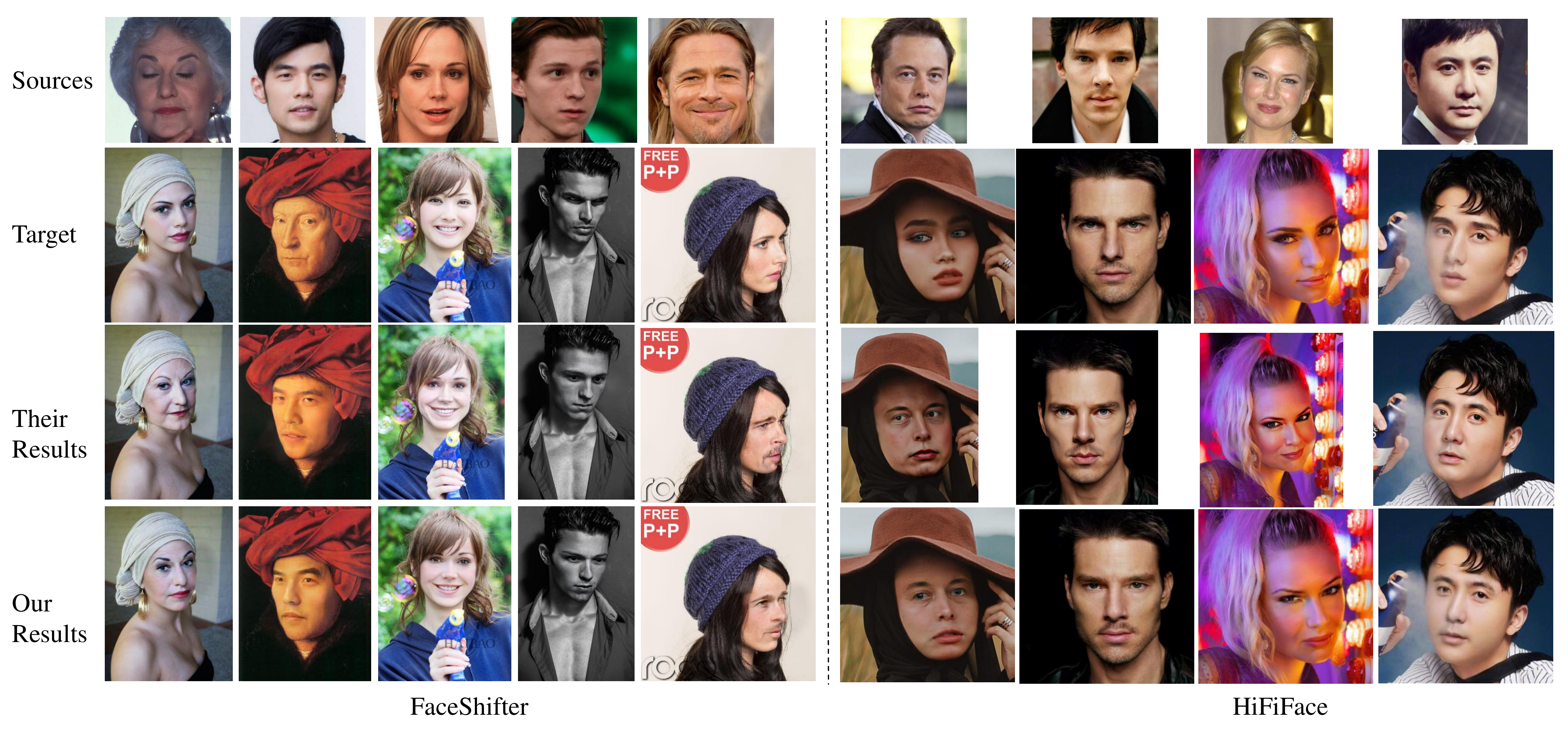}
\centering
\caption{Supplementary qualitative comparisons with FaceShifter and HiFiFace.}
\label{fig:supplementary_comparison}
\end{figure*}

\begin{figure*}[btp]
\includegraphics[width=18cm]{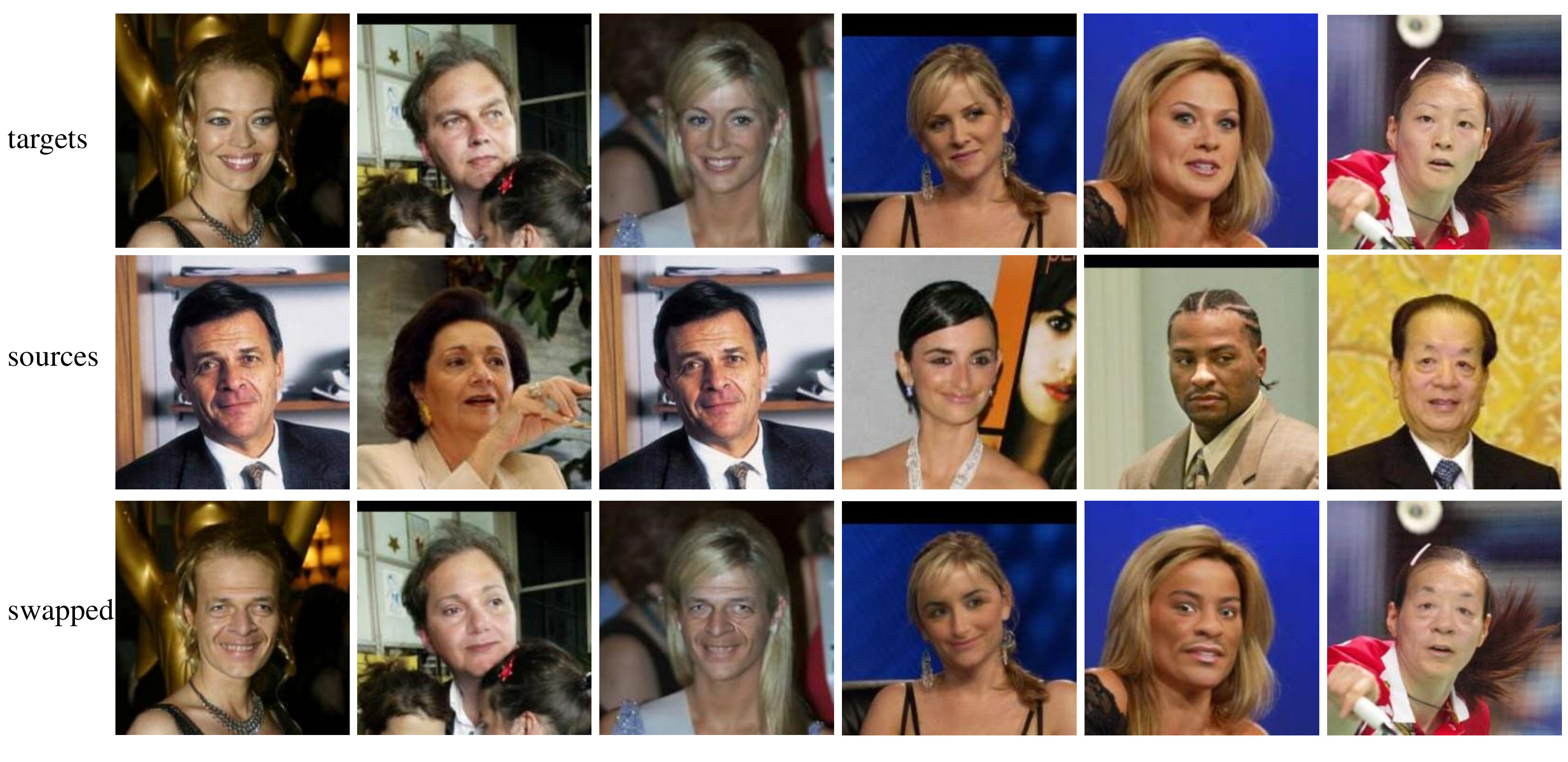}
\centering
\caption{Example face swaps in LFW.}
\label{fig:lfw}
\end{figure*}

\newpage

\section{Appendix II}

In this section, we provide the detailed structures of our proposed generator in figure \ref{fig:encoder} and figure \ref{fig:decoder} as well as their basic building blocks in figures \ref{fig:EncoderHead}, \ref{fig:IdentityEncoderBlock}, \ref{fig:AttributeEncoderBlock} and \ref{fig:DecoderBlock}. 

\begin{figure}[h!bt]
\includegraphics[width=2.8cm]{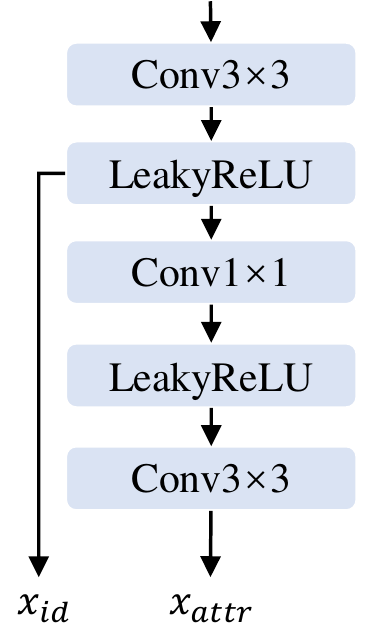}
\centering
\caption{\textmd{\textbf{Encoder Header}. The identity encoder and the attribute encoder share the same header. The encoder header accepts the target image $I_{tgt}$ as input. The first convolution layer converts the RGB image to $N$-channel feature maps. Note that the number of channnels for any feature maps after this convolution layer remains unchanged. The encoder header produces two outputs, each of which serves as the input to the identity encoder branch or the attribute encoder branch. }}
\label{fig:EncoderHead}
\end{figure}

\begin{figure}[h!bt]
\includegraphics[width=2.8cm]{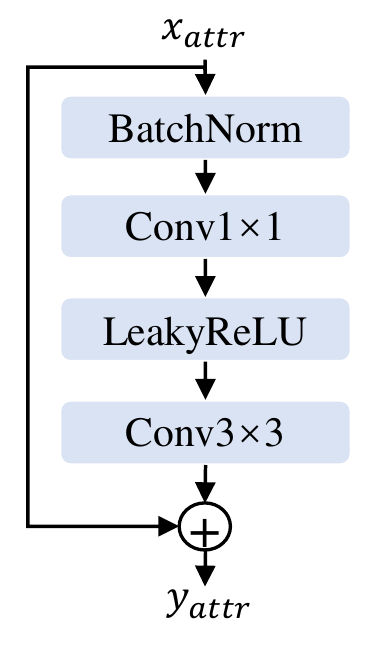}
\centering
\caption{\textmd{\textbf{Attribute Encoder Block}. The attribute encoder blocks preserve the information that can be destroyed by AdaIN. Each of the output feature maps are used in the decoder for the attention module to recover attribute information (e.g. background, hairstyle, expression, pose, etc.). }}
\label{fig:AttributeEncoderBlock}
\end{figure}

\begin{figure}[h!bt]
\includegraphics[width=5.5cm]{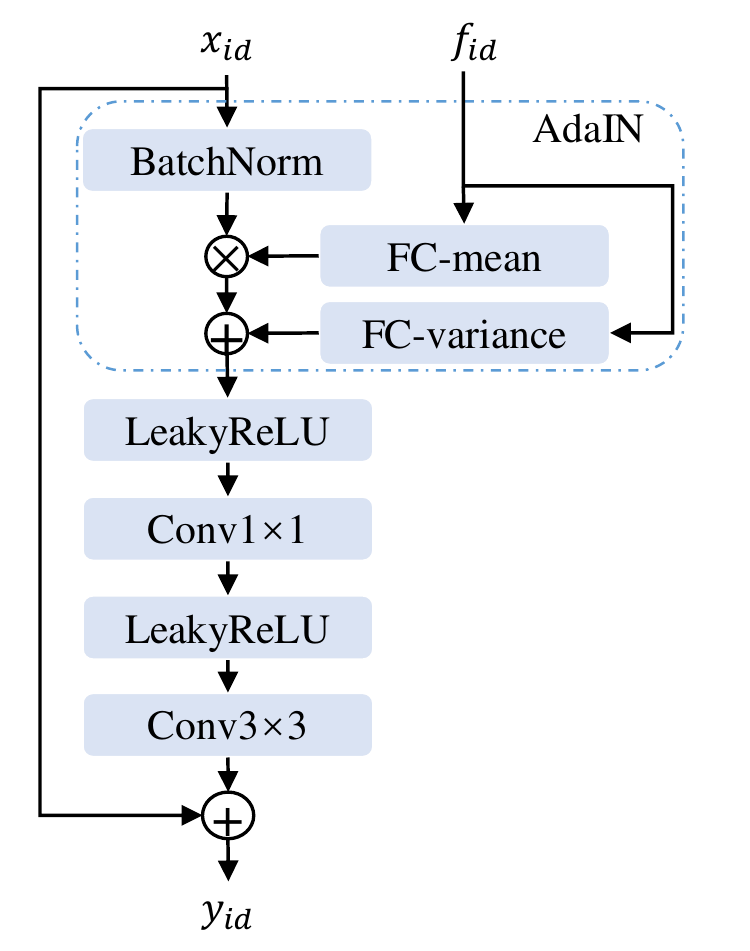}
\centering
\caption{\textmd{\textbf{Identity Encoder Block}. The input feature maps are firstly normalized to a zero-mean unit-variance distribution via the batch normalization layer. Two FC layers learns a new mean and a new variance from the source face embedding $f_{id}$. Then the learned mean and variance are used to re-scale and re-center the distribution. The AdaIN replaces the identity information of the target with that of the source. Then a series of convolutions and nonlinearties are applied. $x_id$ denotes the output from the previous identity encoder block. }}
\label{fig:IdentityEncoderBlock}
\end{figure}

\begin{figure}[h!bt]
\includegraphics[width=3.2cm]{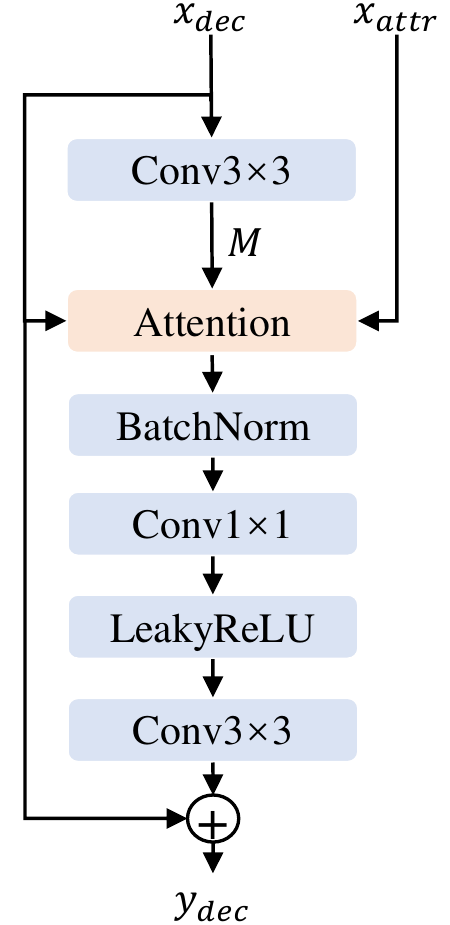}
\centering
\caption{\textmd{\textbf{Decoder Block}. Decoder blocks fuse the source identity information and the attribute information. $x_{dec}$ denotes the output of the previous decoder block. If this is the first decoder block, the $x_{dec}$ is the output of the identity encoder. $x_{attr}$ denotes the output from the corresponding attribute encoder block. The first $3\times 3$ convolution generates an attention map $M$. Then inside the attention module, attribute information and identity information are fused via $y_{att} = M \cdot x_{dec} + (1 - M) \cdot x_{attr}$. }}
\label{fig:DecoderBlock}
\end{figure}

\begin{figure}[h!bt]
\includegraphics[width=9cm]{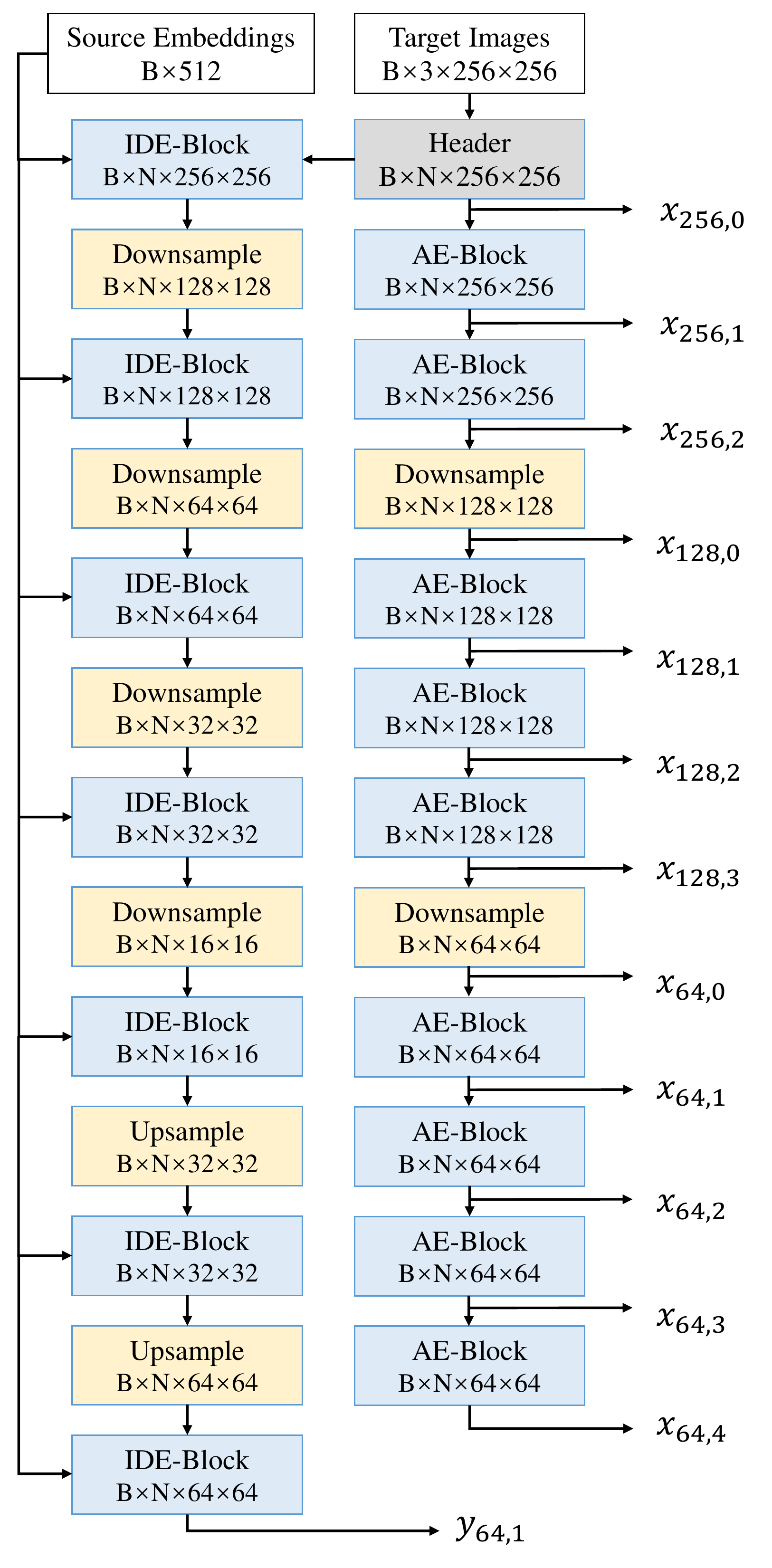}
\centering
\caption{\textbf{Structure of the encoder}. IDE-Block denotes the identity encoder block explained in figure \ref{fig:IdentityEncoderBlock}. AE-Block denotes the attribute encoder block explained in figure \ref{fig:AttributeEncoderBlock}. Structure of the Header is in figure \ref{fig:EncoderHead}.}
\label{fig:encoder}
\end{figure}

\begin{figure}[h!bt]
\includegraphics[width=9cm]{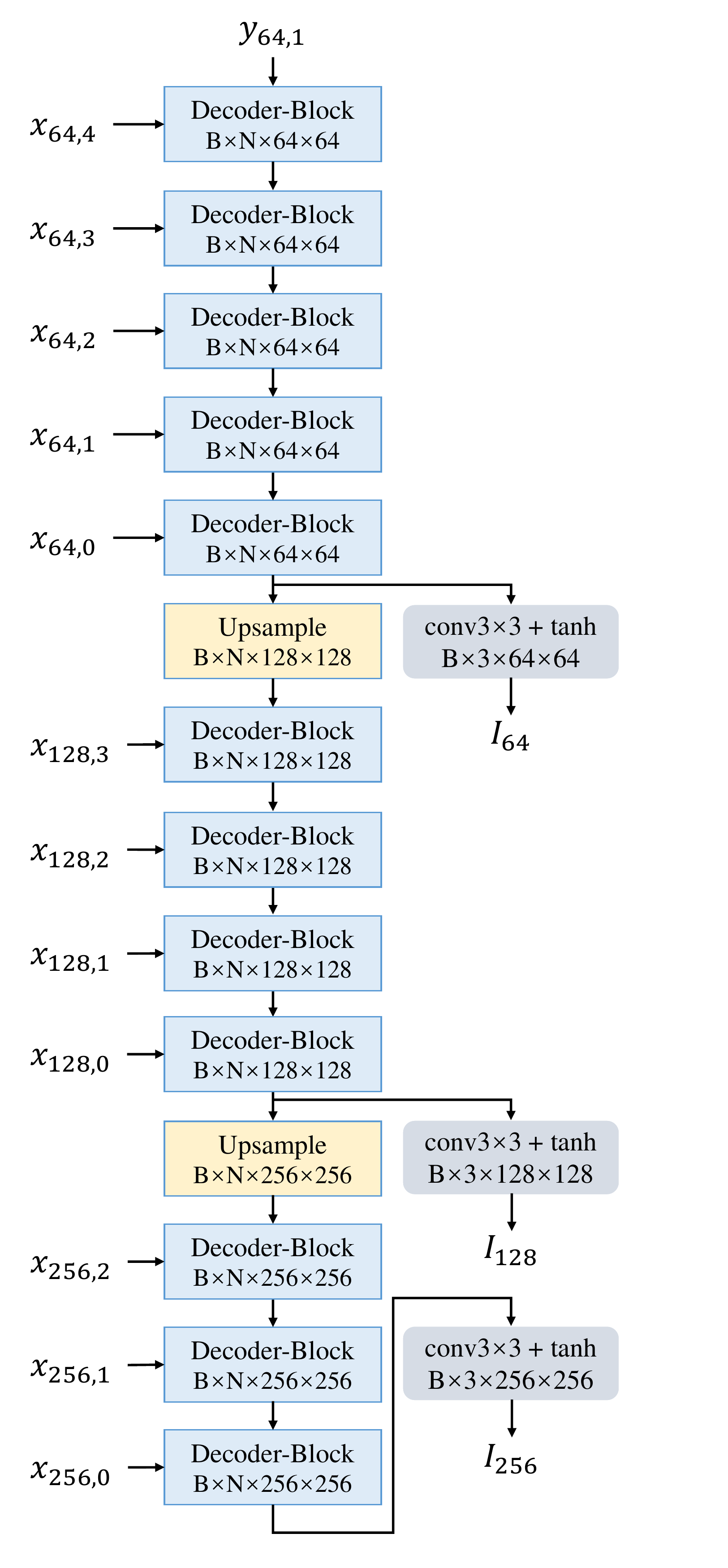}
\centering
\caption{\textbf{Structure of the decoder}. Detailed design of the decoder block is shown in figure \ref{fig:DecoderBlock}. The decoder accepts the feature maps generated by the encoders and produces images of different scales. At the end of each scale, a $3\times 3$ convolution followed by a hyperbolic tangent nonlinearty processes the N-channel feature maps and generates RGB images.}
\label{fig:decoder}
\end{figure}


\sloppy

\def\x{{\mathbf x}}
\def\L{{\cal L}}

\title{Migrating Face Swap to Mobile Devices: A Light-weight Framework and A Supervised Training Solution\protect\\Supplementary Materials}
%
\name{Haiming Yu$^{\ast}$, 
Hao Zhu$^{\dagger}$, 
Xiangju Lu$^{\ast \ddagger}$\thanks{$^{\ddagger}$Corresponding author},
Junhui Liu$^{\ast}$}

\address{$^{\ast}$iQIYI Inc, Beijing, China. \{yuhaiming, luxiangju, liujunhui\}@qiyi.com
\\ $^{\dagger}$Nanjing University, Nanjing, China. zhuhaoese@nju.edu.cn}

\maketitle


\section{Design of the Framework}

Based on our dense experiments, we design our lightweight GAN following these principles:

1, Although increasing the number of channels can enrich the representation power of the feature maps, it results in an exponential growth of the number of parameters. To keep the network as lightweight as possible, we use a fixed number of channels in each layer. We compensate for the loss of representation power by increasing the depth of the network, which introduces new parameters linearly rather than exponentially. 

2, Fusing features from different depths can also enhance the representation ability, as is already demonstrated in DenseNet \cite{densenet} where the feature maps of a layer contain the features from all previous layers. We use residual connections \cite{resnet} for multi-level feature fusion. Meanwhile, we also fuse features in the decoder with those in the encoder by summing the feature maps with the same scale. This not only enhances the richness of the features given such a narrow model but also benefits the updates of the parameters during the training process. 

3, The generator of our face swap model deploys an encoder-decoder structure. Traditionally, an encoder-decoder structure usually follows an hourglass-shaped design, where halving the scale of the feature maps finally results in bottleneck-like features. This restricts the flexibility of designing the architecture. Moreover, spatial information can be lost due to the pooling operation, which is especially important for transferring the identity while preserving the attributes. Instead of such practice, we only down-sample the feature maps three times throughout the whole network, maintaining the size of feature maps unchanged elsewhere. We find that an hourglass-shaped network dramatically attenuates the performance.

Following the above three principles, we designed our framework. We provide the detailed structures of our proposed generator in figure \ref{fig:encoder} and figure \ref{fig:decoder} as well as their basic building blocks in figures \ref{fig:EncoderHead}, \ref{fig:IdentityEncoderBlock}, \ref{fig:AttributeEncoderBlock} and \ref{fig:DecoderBlock}. 

\begin{figure}[bp!]
\includegraphics[width=2.8cm]{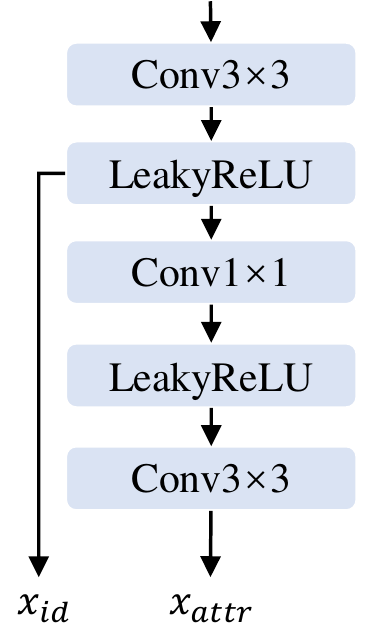}
\centering
\caption{\textmd{\textbf{Encoder Header}. The identity encoder and the attribute encoder share the same header. The encoder header accepts the target image $I_{tgt}$ as input. The first convolution layer converts the RGB image to $N$-channel feature maps. Note that the number of channnels for any feature maps after this convolution layer remains unchanged. The encoder header produces two outputs, each of which serves as the input to the identity encoder branch or the attribute encoder branch. }}
\label{fig:EncoderHead}
\end{figure}

\begin{figure}[bp!]
\includegraphics[width=2.8cm]{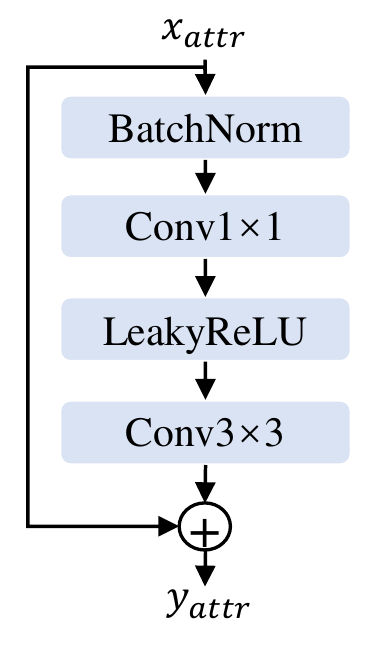}
\centering
\caption{\textmd{\textbf{Attribute Encoder Block}. The attribute encoder blocks preserve the information that can be destroyed by AdaIN. Each of the output feature maps are used in the decoder for the attention module to recover attribute information (e.g. background, hairstyle, expression, pose, etc.). }}
\label{fig:AttributeEncoderBlock}
\end{figure}

\begin{figure}[bp!]
\includegraphics[width=5.5cm]{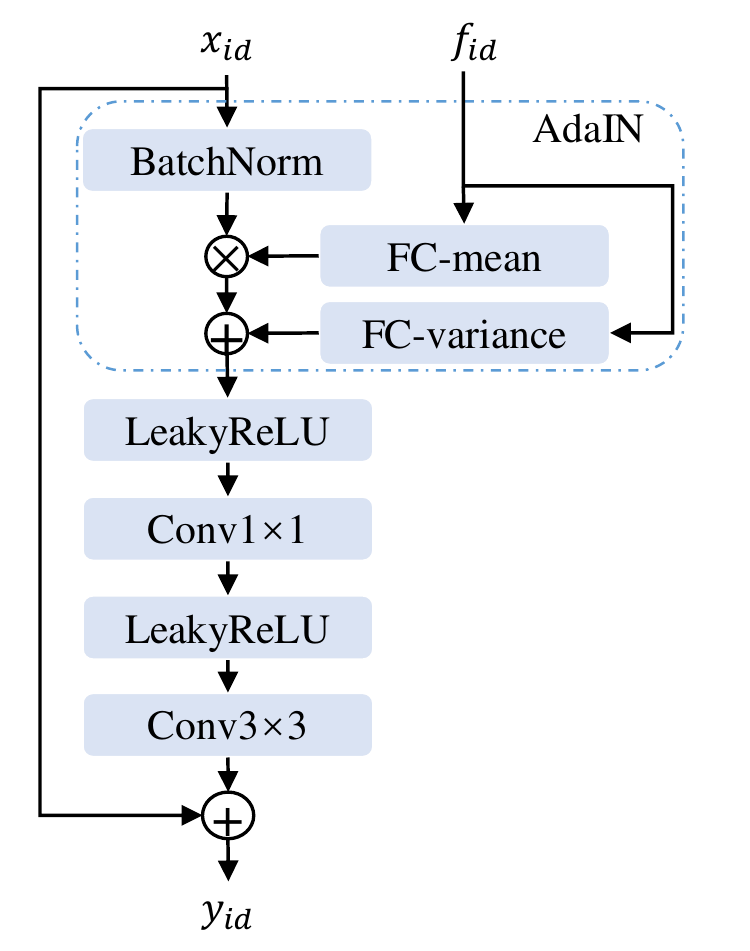}
\centering
\caption{\textmd{\textbf{Identity Encoder Block}. The input feature maps are firstly normalized to a zero-mean unit-variance distribution via the batch normalization layer. Two FC layers learns a new mean and a new variance from the source face embedding $f_{id}$. Then the learned mean and variance are used to re-scale and re-center the distribution. The AdaIN replaces the identity information of the target with that of the source. Then a series of convolutions and nonlinearties are applied. $x_id$ denotes the output from the previous identity encoder block. }}
\label{fig:IdentityEncoderBlock}
\end{figure}

\begin{figure}[bp!]
\includegraphics[width=3.2cm]{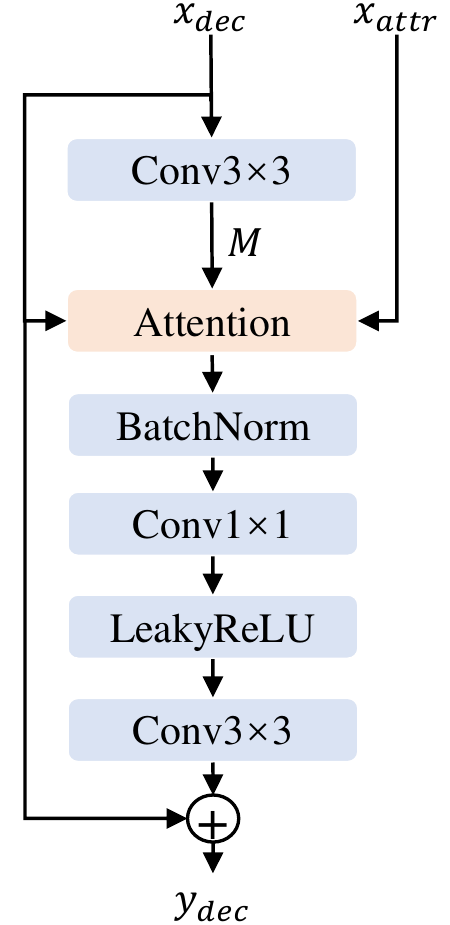}
\centering
\caption{\textmd{\textbf{Decoder Block}. Decoder blocks fuse the source identity information and the attribute information. $x_{dec}$ denotes the output of the previous decoder block. If this is the first decoder block, the $x_{dec}$ is the output of the identity encoder. $x_{attr}$ denotes the output from the corresponding attribute encoder block. The first $3\times 3$ convolution generates an attention map $M$. Then inside the attention module, attribute information and identity information are fused via $y_{att} = M \cdot x_{dec} + (1 - M) \cdot x_{attr}$. }}
\label{fig:DecoderBlock}
\end{figure}

\begin{figure}[bp!]
\includegraphics[width=9cm]{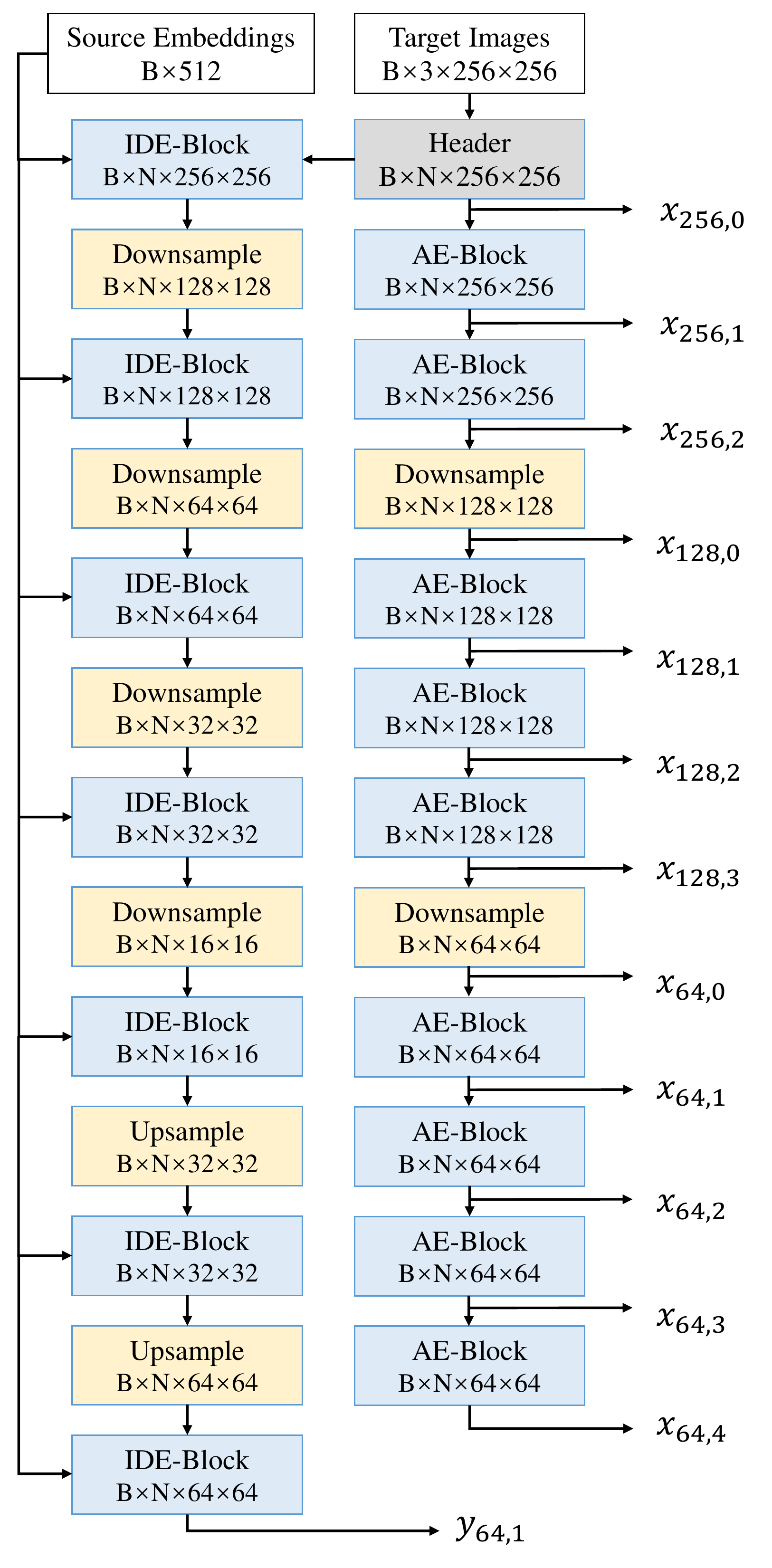}
\centering
\caption{\textbf{Structure of the encoder}. IDE-Block denotes the identity encoder block explained in figure \ref{fig:IdentityEncoderBlock}. AE-Block denotes the attribute encoder block explained in figure \ref{fig:AttributeEncoderBlock}. Structure of the Header is in figure \ref{fig:EncoderHead}.}
\label{fig:encoder}
\end{figure}

\begin{figure}[bp!]
\includegraphics[width=9cm]{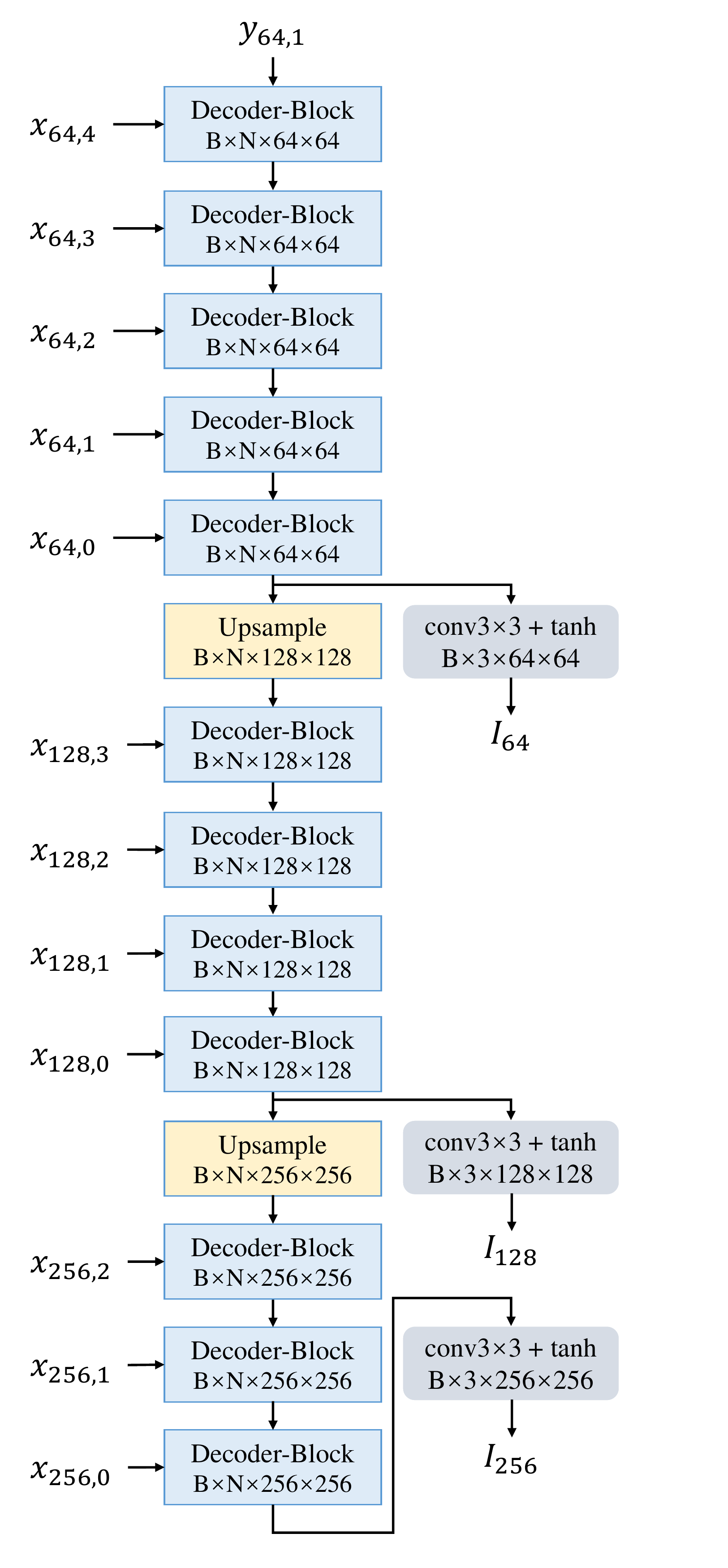}
\centering
\caption{\textbf{Structure of the decoder}. Detailed design of the decoder block is shown in figure \ref{fig:DecoderBlock}. The decoder accepts the feature maps generated by the encoders and produces images of different scales. At the end of each scale, a $3\times 3$ convolution followed by a hyperbolic tangent nonlinearty processes the N-channel feature maps and generates RGB images.}
\label{fig:decoder}
\end{figure}

\begin{table}[]
\centering
\caption{Results on FaceForensics++ for ablation study on network design. }
\label{tab:ablation_study}
\begin{tabular}{l|ccccc}
 & \multicolumn{1}{l}{ID} & \multicolumn{1}{l}{Pose} & \multicolumn{1}{l}{Exp.} & \multicolumn{1}{l}{FID} & \multicolumn{1}{l}{Size/MB} \\ \hline
Wide     & 99.48 & 2.32 & 2.84 & 10.17 & 586.0 \\
Shallow & 94.86 & 1.95 & 2.67 & 9.75  & 8.8   \\
NoFuse    & 97.84     & 15.80    & 3.65    & 277.01     & 10.2     \\
HG & 34.40 & 3.60 & 2.52 & 39.10 & 12.4  \\ \hline
Baseline     & 98.48 & 2.18 & 2.68 & 9.54  & 10.2 
\end{tabular}
\end{table}

\section{Ablation Study on Network Design}

In this section, we conduct several experiments to validate three principles for designing our network, illustrating how these settings affect the performance.  We train the four models and test their performance on FaceForensics++. Besides the three aforementioned metrics, we also report the Frechet Inception Distance (FID) \cite{fidscore} and the model size. We denote our model used in previous sections as \textbf{baseline}. Other models include:

\textbf{WiderNet (Wide)}: Since we have claimed that expanding the width of feature maps as the network goes deeper provides very limited improvement while introducing new parameters exponentially, we design a network with identical structures except that we double the number of channels after each down-sampling layer and halve the number of channels after each up-sampling layer. 

\textbf{ShallowerNet (Shallow)}: We demonstrate the importance of depth by designing a shallower network. For each scale of the feature maps, we remove one basic building block from the attribute encoder and the decoder. Others remain unchanged.

\textbf{NonFusingNet (Nofuse)}: To demonstrate the effectiveness of feature fusion, we remove all connections for feature fusing while keeping other settings unchanged.

\textbf{HourglassNet (HG)}: We use the same basic building blocks to build an hourglass-shaped generator, down-sampling the feature maps until it becomes a vector at the end of the encoders. The depth of this network remains the same as that of the baseline model.

The results are reported in table \ref{tab:ablation_study}. By comparing the results, we draw the following conclusions:

\textbf{Depth is preferable to width}: Comparing WiderNet with the baseline, we can witness that expanding the width of the network does not show obvious superiority while introducing new parameters exponentially. However, as is shown in ShallowerNet, removing only a few layers already linearly degrade the performance. Therefore, a deeper network is preferable to a wider network. 

\textbf{Feature Fusion enriches the representation ability}: The accuracy of ID retrieval remains high for NonFusingNet. However, PID indicates that the image quality drops sharply. In fact, this model collapsed to a point where unrealistic images are produced while the identity-preserving loss is very low, indicating the model failed to balance both goals. 

\textbf{Hourglass-shaped encoder-decoder structure imposes a bottleneck for information flow}: We can notice a sharp performance drop in the results of HourglassNet. This is due to the fact that spatial information is lost if the scale of the feature maps is too small, imposing an obstacle for spatially fusing the identity and attribute information. Hence the network fails to transfer the identity and the image quality drops.

\section{More Qualitative results}

Lastly, we provide more qualitative comparisons with FaceShifter \cite{faceshifter} and HiFiFace \cite{hififace}. We also exhibit results from experiments on LFW \cite{lfw}.

\begin{figure*}[bp!]
\includegraphics[width=18cm]{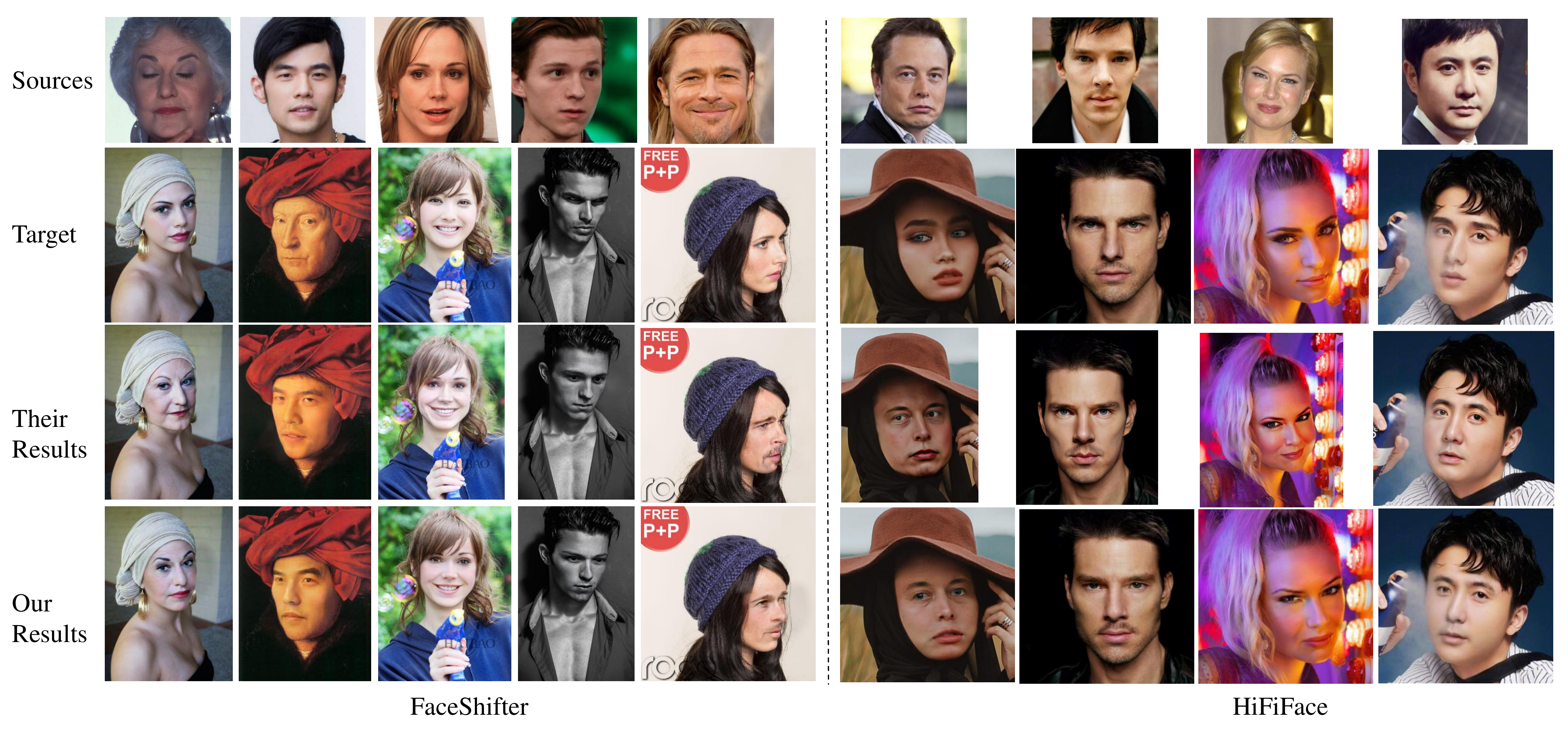}
\caption{Supplementary qualitative comparisons with FaceShifter and HiFiFace.}
\label{fig:supplementary_comparison}
\end{figure*}

\begin{figure*}[bp!]
\includegraphics[width=18cm]{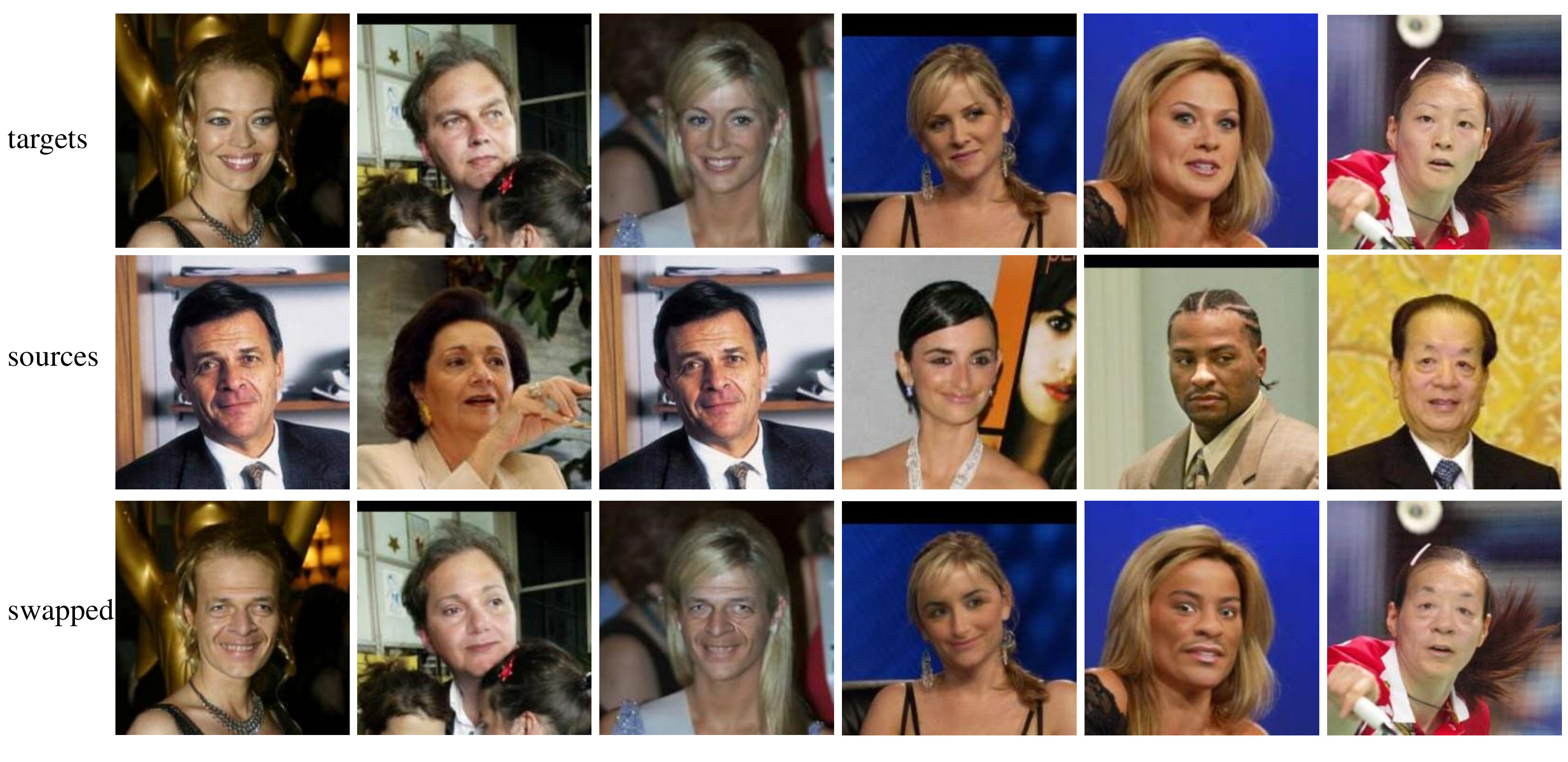}
\centering
\caption{Example face swaps in LFW.}
\label{fig:lfw}
\end{figure*}

\bibliographystyle{IEEEbib}
\bibliography{supp}